\documentclass[letterpaper, 10 pt, conference]{ieeeconf}  

\IEEEoverridecommandlockouts
\usepackage{cite}
\usepackage{algorithm}
\def\BibTeX{{\rm B\kern-.05em{\sc i\kern-.025em b}\kern-.08em
    T\kern-.1667em\lower.7ex\hbox{E}\kern-.125emX}}

\usepackage[utf8]{inputenc}
\usepackage[english]{babel}

\usepackage{amsmath} 
\usepackage{amssymb}
\usepackage{amsfonts}
\usepackage{mathtools}

\usepackage{bm}
\usepackage{cite}
\usepackage{comment}
\allowdisplaybreaks 

\usepackage[pdftex]{graphicx} 
\usepackage{epstopdf}
\usepackage{hyperref} 
\usepackage{color}
\usepackage{xcolor}
\definecolor{Questions}{HTML}{1F77B4}
\usepackage[font=footnotesize,labelfont=bf]{caption}
\usepackage{lipsum}
\usepackage{subcaption}

\usepackage{tikz}
\usetikzlibrary{calc}
\usetikzlibrary{arrows,positioning} 
\usetikzlibrary{intersections, patterns}
\usetikzlibrary{shapes}
\usepackage{textcomp}
\usepackage{mathtools}
\usepackage{graphicx}
\usepackage{bbm}
\usepackage[cmtip,all]{xy}
\newcommand{\longsquiggly}{\xymatrix{{}\ar@{~>}[r]&{}}}
\usepackage{algpseudocode}

\algnewcommand{\Inputs}[1]{%
  \State \textbf{Inputs:}\hspace*{\algorithmicindent}\parbox[t]{.8\linewidth}{\raggedright #1}
}
\algnewcommand{\Initialize}[1]{%
  \State \textbf{Initialize:}\hspace*{\algorithmicindent}\parbox[t]{.8\linewidth}{\raggedright #1}
}

\begin{document}
\title{Quad-LCD: Layered Control Decomposition Enables Actuator-Feasible Quadrotor Trajectory Planning}

\author{Anusha Srikanthan, Hanli Zhang, Spencer Folk, Vijay Kumar, Nikolai Matni
\thanks{We gratefully acknowledge the support of NSF Grant CCR-2112665, NSF awards CPS-2038873, SLES-2331880, AFOSR Award FA9550-24-1-0102 and NSF CAREER award ECCS-2045834. for this research. 
}
\thanks{A. Srikanthan, S. Folk, V. Kumar, N. Matni are with the School of Engineering and Applied Science, University of Pennsylvania, Philadelphia, USA (e-mail:\{sanusha, sfolk, kumar, nmatni\}@seas.upenn.edu) and H. Zhang is with EPFL, Lausanne.}}

\maketitle

\begin{abstract}
    In this work, we specialize contributions from prior work on data-driven trajectory generation for a quadrotor system with motor saturation constraints.
    When motors saturate in quadrotor systems, there is an ``uncontrolled drift" of the vehicle that results in a crash. To tackle saturation, we apply a control decomposition and learn a tracking penalty from simulation data consisting of low, medium and high-cost reference trajectories. Our approach reduces crash rates by around $49\%$ compared to baselines on aggressive maneuvers in simulation. On the Crazyflie hardware platform, we demonstrate feasibility through experiments that lead to successful flights. Motivated by the growing interest in data-driven methods to quadrotor planning, we provide open-source lightweight code with an easy-to-use abstraction of hardware platforms. 
\end{abstract}


\section{Introduction}

Recently, quadrotor tracking control has garnered the attention from deep learning communities, most notably culminating in works like Neural-fly~\cite{connell2022neural} and DATT~\cite{huang2023datt}. In~\cite{connell2022neural}, the authors train neural network control policies to compensate for aerodynamic wrenches based on as little as 12 minutes of flight data and in~\cite{huang2023datt}, the authors demonstrate improved trajectory tracking for feasible and infeasible planar trajectories in $(x, y)$ for a fixed altitude. Other approaches that use deep learning methods~\cite{hwangbo2017control, zhang2023learning, molchanov2019sim} primarily study the problem of quadrotor stabilization and sim-to-real policy transfer. However, note that all the approaches discussed so far directly learn feedback policies to account for external perturbations.  

We consider a quadrotor system with a fixed nonlinear feedback tracking controller. Differing from prior work on adaptive control for quadrotor systems~\cite{svacha2017improving, wu20221, huang2023datt, sanghvi2024occam}, we use simulation data to improve existing trajectory planners to adapt to the fixed controller's capabilities. Following prior work~\cite{srikanthan2023data}, we apply a layered control decomposition to obtain a controller-aware planning problem that optimizes reference trajectories. 
By doing so, we reverse engineer to obtain a trajectory generation problem that includes a tracking penalty. 
This tracking penalty learned from trajectory data implicitly avoids aggressive behavior by reshaping paths between waypoints. We demonstrate our data-driven planner's effectiveness by testing on various quadrotor parameters such as drag coefficients and provide a streamlined training and inference pipeline. 
``Quad-LCD"~\footnote{\url{https://github.com/Nusha97/Quad-LCD}} is, to our knowledge, the first open-source, lightweight abstraction of quadrotor simulation capabilities for training and deploying data-driven planners. Therefore, our contributions are as follows:


\begin{enumerate}
\item We provide large-scale, parallelizable data collection of low, medium, and high cost long-horizon reference trajectories, including infeasible trajectories via a realistic Python-based simulator;
\item We demonstrate significant reduction in crash rates, and waypoint tracking error on a Crazyflie platform;
\item Our open-source implementation also provide a ROS Python interface for hardware deployment and enables opportunities for careful design of data-driven methods with the possible integration of estimation, vision and language in realistic simulators.
\end{enumerate}

\begin{figure}[t]
    \centering
    \includegraphics[width=\columnwidth]{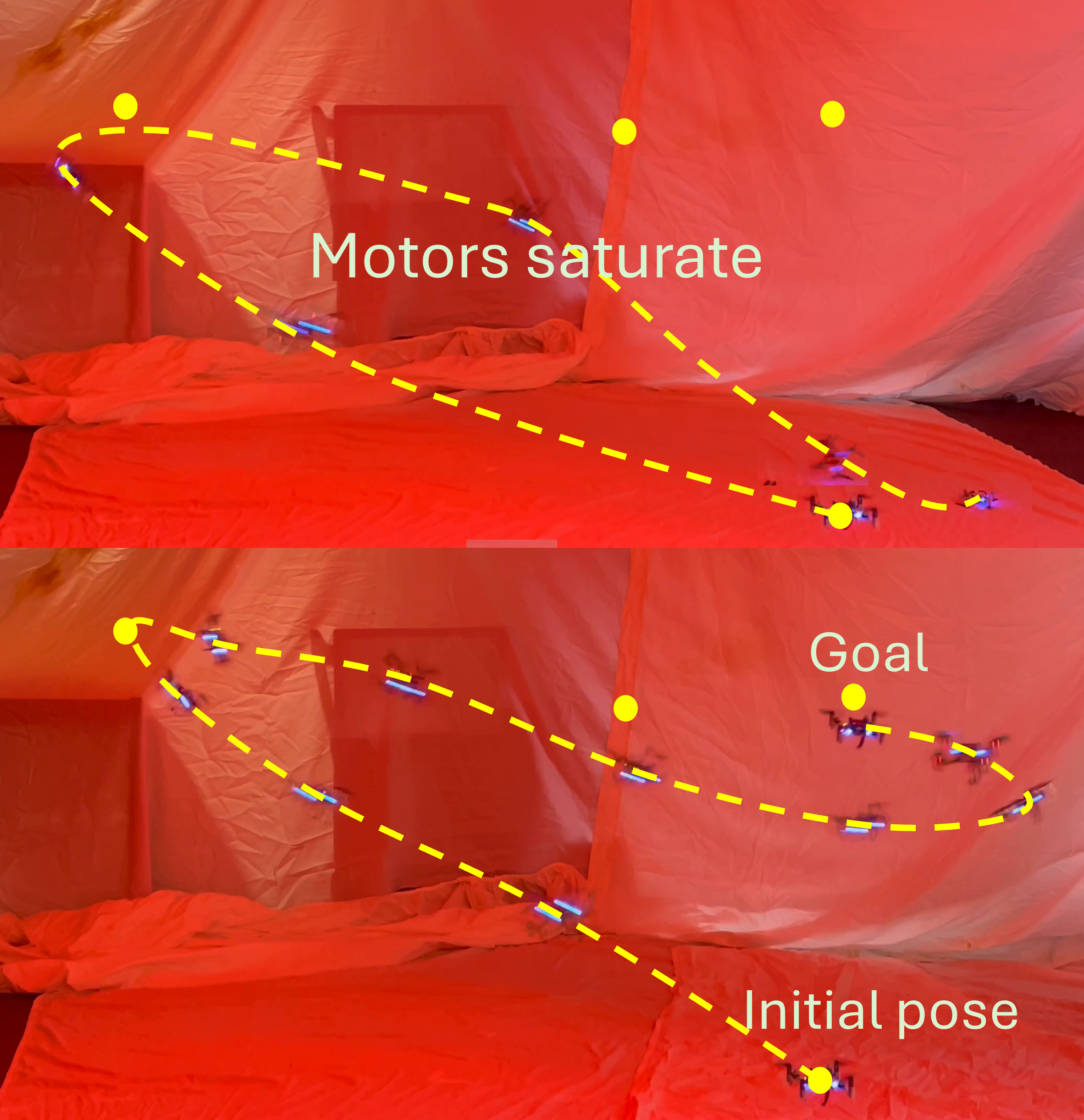}
    \caption{We show the flight path of a Crazyflie 2.0 resulting in a crash due to motor saturation from aggressive flight on the top and successful flight below. The reference paths were designed using a baseline planner~\cite{mellinger2011minimum} and our approach, respectively.}
    \label{fig:quad-hw}
\end{figure}

\section{Problem Formulation}

We adopt the notation of nonlinear dynamics of a quadrotor from~\cite{RichterBryRoy2016} given by
\begin{equation}
    \begin{aligned}
        m\ddot{\boldsymbol{r}} &= m g \boldsymbol{z}_W - f \boldsymbol{z}_B, \\
        \dot{\omega} &= J^{-1} [-\omega \times J \omega + M]
    \end{aligned}
\end{equation}
where $\boldsymbol{r}$ is the position in the world-coordinate frame defined by the unit vector $\boldsymbol{z}_W$ along the direction of gravity, $\omega$ is the angular velocity in the body-fixed coordinate frame, $f$ and $M$ are net forces and moments, respectively defined in the body frame (denoted by unit vector $\boldsymbol{z}_B$). $J$ and $m$ correspond to the inertial moment tensor and mass of the vehicle respectively. The desired motor speeds are obtained as a function of individual rotor thrusts allocated based on the desired net thrusts and moments along each axis. 

We fix a nonlinear feedback controller from~\cite{lee2010geometric} given by equations for $f$ and $M$ as
\begin{equation}\label{eq:nonlinear-ctrl}
    \begin{aligned}
        f &= (-k_x \boldsymbol{e}_x - k_v \boldsymbol{e}_v + m g \boldsymbol{z}_W + m \ddot{\boldsymbol{r}}_d) \cdot R \boldsymbol{z}_W, \\
        M &= -k_R \boldsymbol{e}_R - k_{\omega} \boldsymbol{e}_{\omega} + \omega \times J \omega - J(\hat{\omega}R^TR_d\omega_d - R^TR_d \dot{\omega})
    \end{aligned}
\end{equation}
where $R$ is a rotation matrix, $\boldsymbol{e}_R, \boldsymbol{e}_x, \boldsymbol{e}_v, \boldsymbol{e}_{\omega}$ are the errors in rotation, position, speed and angular speed, $k_R, k_x, k_v, k_{\omega}$ are corresponding control gains. A desired trajectory $r_d$ and orientation $R_d$ is specified by independent polynomial functions of time of position and yaw angles $(x, y, z, \psi)$. The derivatives of the polynomials are sufficient to determine the control inputs in equations~\eqref{eq:nonlinear-ctrl} at time $t$ eliminating the need for numerical integration which is a powerful tool for simulation. This simulation capability is exploited in RotorPy~\cite{folk2023rotorpy} which we use as our simulator in this work. 

Although differential flatness provides this powerful capability, there is an inherent assumption made that the angular speeds $\omega$ in the body frame are almost equal to their values in the world frame. This assumption is valid for quadrotors that are close to hover or do not roll, pitch or yaw aggressively. Further, actuator constraints on motor speeds are unaccounted for and as discussed in~\cite{RichterBryRoy2016}, they are typically handled by reformulating the problem to find a better time allocation. There have been several works~\cite{verscheure2009time, gao2018optimal} that discuss methods to parameterize a time-optimal path. In this paper, we ask a slightly different question. \emph{Given a fixed nonlinear controller as in~\eqref{eq:nonlinear-ctrl} and a simulator, how do we use simulation data to optimize for actuator-feasible reference trajectories?}

\section{Optimizing for Actuator-Feasible Polynomial Trajectories}\label{sec:actuator-poly}

As discussed before, the assumption of differential flatness for simulation is valid and enables zero-shot deployment without sim-to-real gap as long as trajectories are actuator-feasible. What we mean by actuator-feasible is that the commanded motor speeds obtained from the nonlinear feedback controller lies in the operating range of motor capacities. Aggressive trajectories that result in unmodeled dynamics that cause rolling, pitching or yawing of the quadrotor at high speeds violate this assumption. Instead of modeling residual dynamics directly, we take an approach inspired by~\cite{srikanthan2023data}. The cost functional for minimizing snap associated with optimizing the $i$th polynomial segment~\cite{mellinger2011minimum, RichterBryRoy2016} is given by
\begin{equation*}
    J_i(T) = \int_{0}^T \| \ddddot{x}(t) \|_2^2 dt = \boldsymbol{c}_i^T H_i \boldsymbol{c}_i
\end{equation*}
where $c_i$ is an $n$th order polynomial and $H_i$ is the cost function obtained by differentiating through the polynomials. Concatenating $s$ such polynomial segments and incorporating a controller-aware cost functional~\cite{srikanthan2023data}, we obtain the optimization problem as
\begin{equation}\label{prob:drag-aware-plan}
    \begin{array}{rl}
         \underset{\boldsymbol{c}}{\mathrm{minimize}}& \boldsymbol{c}^T H \boldsymbol{c} + g^{ctrl}(\xi, \boldsymbol{c})  \
         \text{subject to}
         \, A \boldsymbol{c} = b
    \end{array}
\end{equation}
where $\boldsymbol{c} \in \mathbb{R}^{4s(n+1)}$ represents the stacked coefficients of piece-wise polynomials of order $n$ with $s$ segments, $H$ is a block diagonal matrix consisting of $H_i$ over $s$ segments, $A, b$ define continuity and smoothness constraints at segment end points on position, yaw and its higher order derivatives up to jerk, and $\xi$ the initial state. For a particular set of coefficients $\Bar{c}$, $g^{ctrl}(\xi, \Bar{c})$ maps $\Bar{c}$ to the sum of tracking error deviations between the commanded and executed trajectories. In simulation, we model residual effects of aggressive motion by setting the motor response time as $5$ milliseconds, and adding motor noise with a standard deviation of $100$ radians per second. In the following section, we address the issue of computing $g^{ctrl}(\xi, \boldsymbol{c})$.

\section{Learning tracking cost from simulation data}

The tracking cost function $g^{ctrl}(\xi, \boldsymbol{c})$ represents the ability of the controller in~\eqref{eq:nonlinear-ctrl} to track reference states also known as its~\emph{value function}. Finding analytic expressions for the value function may be possible and there has been recent work~\cite{teng2022lie} that study cost function design on Lie groups. To avoid having to explicitly model effects of motor noise for cost function design on Lie groups, we use supervised learning to implicitly learn a map from polynomial coefficients to tracking cost. The learned value function acts as a regularizer in the planning problem~\eqref{prob:drag-aware-plan} biasing polynomial coefficients toward actuator-feasible solutions which we will demonstrate through experiments. Learning the tracking cost from simulation data enables testing the effects of varying controller or environment parameters that may not otherwise yield clean mathematical equations for analysis.

\textbf{Remark}: In prior work~\cite{srikanthan2023data}, the authors apply a supervised learning approach directly on an augmented dynamical system resulting in the linear scaling of the input dimension with the rate of discretization frequency. As controllers are typically run at $100$ Hz or $1000$ Hz, the problem requires large amounts of data to train the network and quickly becomes intractable. For example, the nonlinear tracking controller~\eqref{eq:nonlinear-ctrl} running at $100$ Hz will require an input dimension of $1700$. In our approach, we avoid this by using a tractable polynomial representation of fixed order $n$ which scales with the number of segments $s$.




\vspace{-8pt}

\section{Simulation and Experiments}\label{sec:simulation}

To demonstrate our method, in addition to motor noise we also vary the parasitic drag coefficients to showcase the benefits of learning from simulation data. We choose $5$ values of drag coefficients by symmetrically scaling $x, y, z$ components of the parasitic drag matrix defined in~\cite[Sec. II.B]{folk2023rotorpy} from a range over $[0.002, 0.008]$ N/(m/s) for $x, y$ and $[0.007, 0.013]$ N/(m/s) for $z$. For each experiment configuration, we collect data and train a separate network, and evaluate each network's performance on $100$ waypoint-following tasks. We consider three baselines adapted from prior work~\cite{huang2023datt, svacha2017improving, mellinger2011minimum} to compare with i) a class of approaches that solves via end-to-end RL termed ``deep-RL"; ii) minimum snap trajectory generation with a $SE(3)$ geometric controller termed ``MS-GC" and iii) minimum snap trajectory generation with an adaptive controller for drag compensation termed ``MS-GCD". For our RL policy, we implement a custom version of DATT~\cite{huang2023datt} and train our policy using curriculum learning by fixing a reference trajectory seed for the first $2.5$ million steps and vary the seed after every $50,000$ steps training for a total of $10$ million steps. 


\textbf{Data Collection}\label{sec:data-collection}: We collected rollouts of the nonlinear controller in~\eqref{eq:nonlinear-ctrl} on a total of $200,000$ minimum snap trajectories using RotorPy. Each trajectory was generated by random sampling of four waypoints in sequence in a $10\times10\times10\:m^3$ domain, ensuring that each waypoint was at least $1\:m$ and no more than $3\:m$ apart from the previous one. For each set of waypoints, we planned a minimum snap trajectory using $v_{avg}=2\:m/s$ as a heuristic for time allocation. During each rollout, the cumulative position and yaw tracking error was recorded as an estimate of tracking penalty for the coefficients. The minimum snap coefficients and the corresponding tracking penalty from the rollout constitute the labeled pairs used for supervised learning. By leveraging parallelization over 40 CPU cores available on a {\tt 2x AMD EPYC 9684X 96-Core Processor}, we were able to simulate hundreds of hours of quadrotor trajectory rollouts in just 6 hours. We leave further optimizations to parallelize over GPUs as future work.

\begin{figure}[t]
    \centering
    \includegraphics[width=\columnwidth]{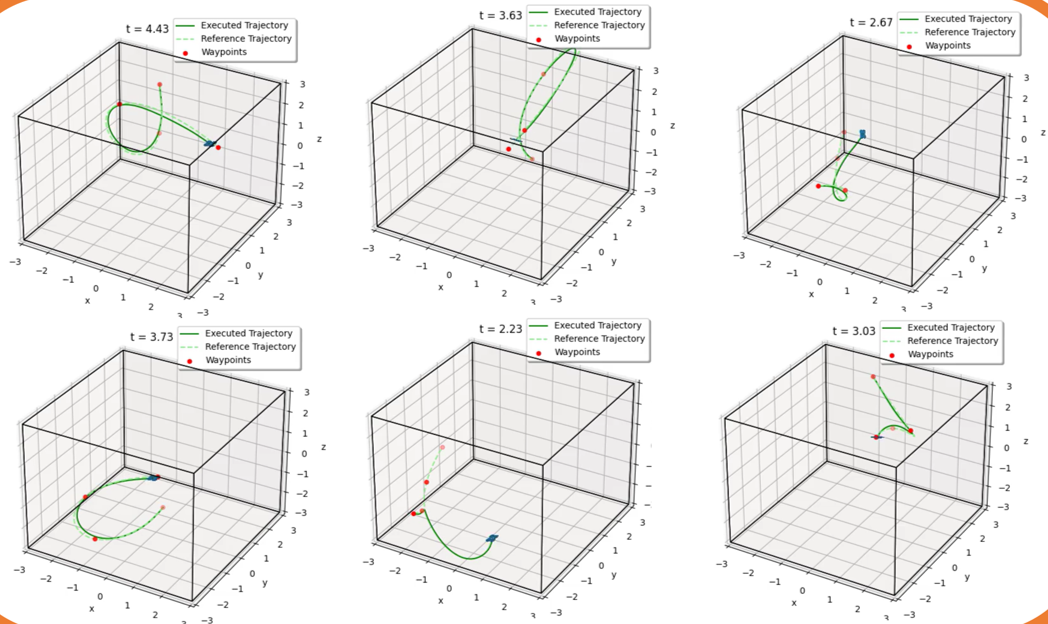}
    \caption{A visualization of low, medium and high-cost trajectories simulated on RotorPy.}
    \vspace{-10pt}\label{fig:trajectoies}
\end{figure}

\begin{figure*}
    \centering
    \includegraphics[width=\textwidth, height=2.8in]{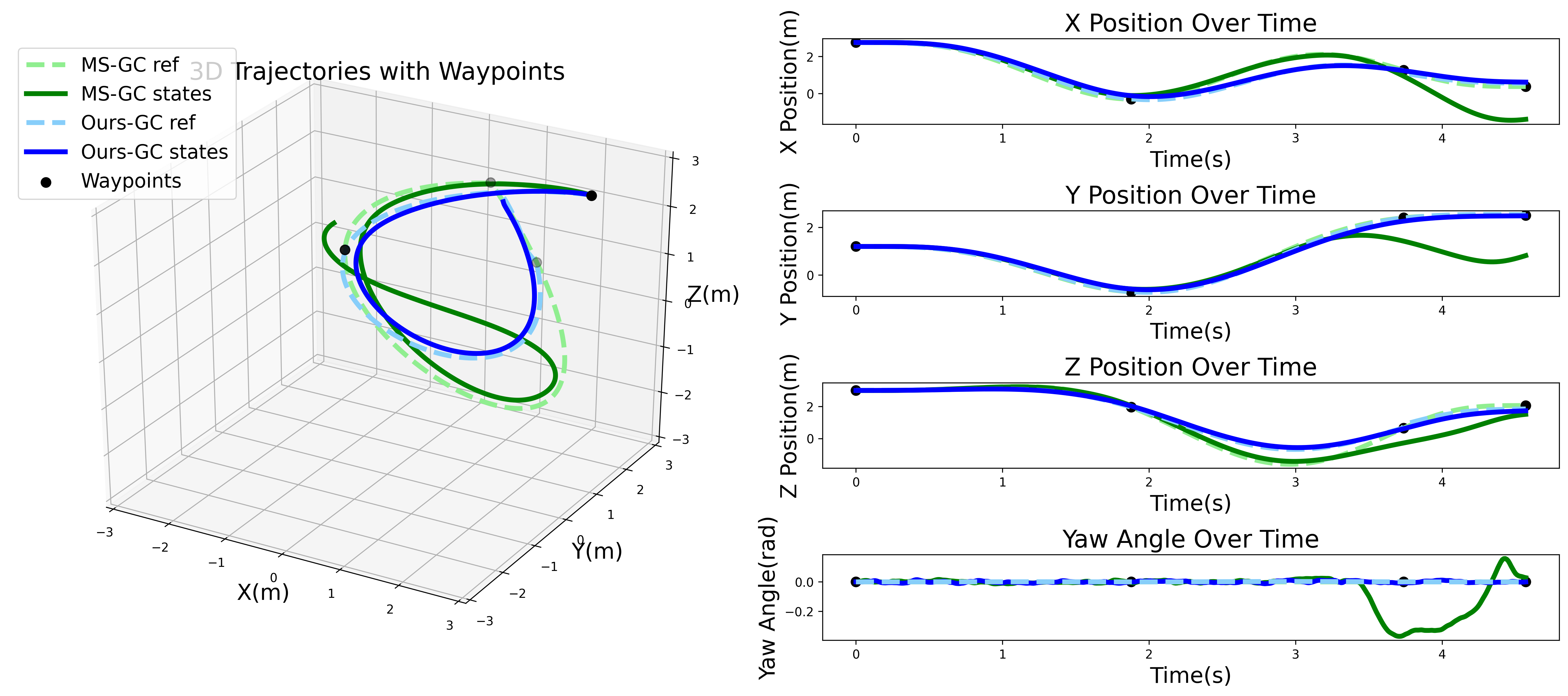}
    \caption{We show a visualization of trajectories simulated on RotorPy for the standard Crazyflie platform where dotted and solid lines are reference and controller executed trajectories, respectively. On the left is a 3D plot showing the deviation of controller executed trajectories and reference for our approach and a baseline. On the right, we plot the $x, y, z$ and $\psi$ curves with waypoints. The deviations in $x$ and $y$ for the baseline planner is high due to large swings from controller saturation while Ours-GC plans references that are tracked more accurately.}
    \vspace{-10pt}
    \label{fig:3d-plots}
\end{figure*}

\textbf{Training and Cross-Validation}: We split the collected data into 80\% training and 20\% validation sets and train a separate multi-layer perceptron (MLP) network with $3$ hidden layers of $\{100, 100, 20\}$ neurons, respectively, with Rectified Linear Unit (ReLU) activation functions for each experiment configuration. Further evaluation with input convex neural networks~\cite{amos2017input} are left as future work.

\begin{figure}[t]
        \centering
        \includegraphics[width=1\columnwidth]{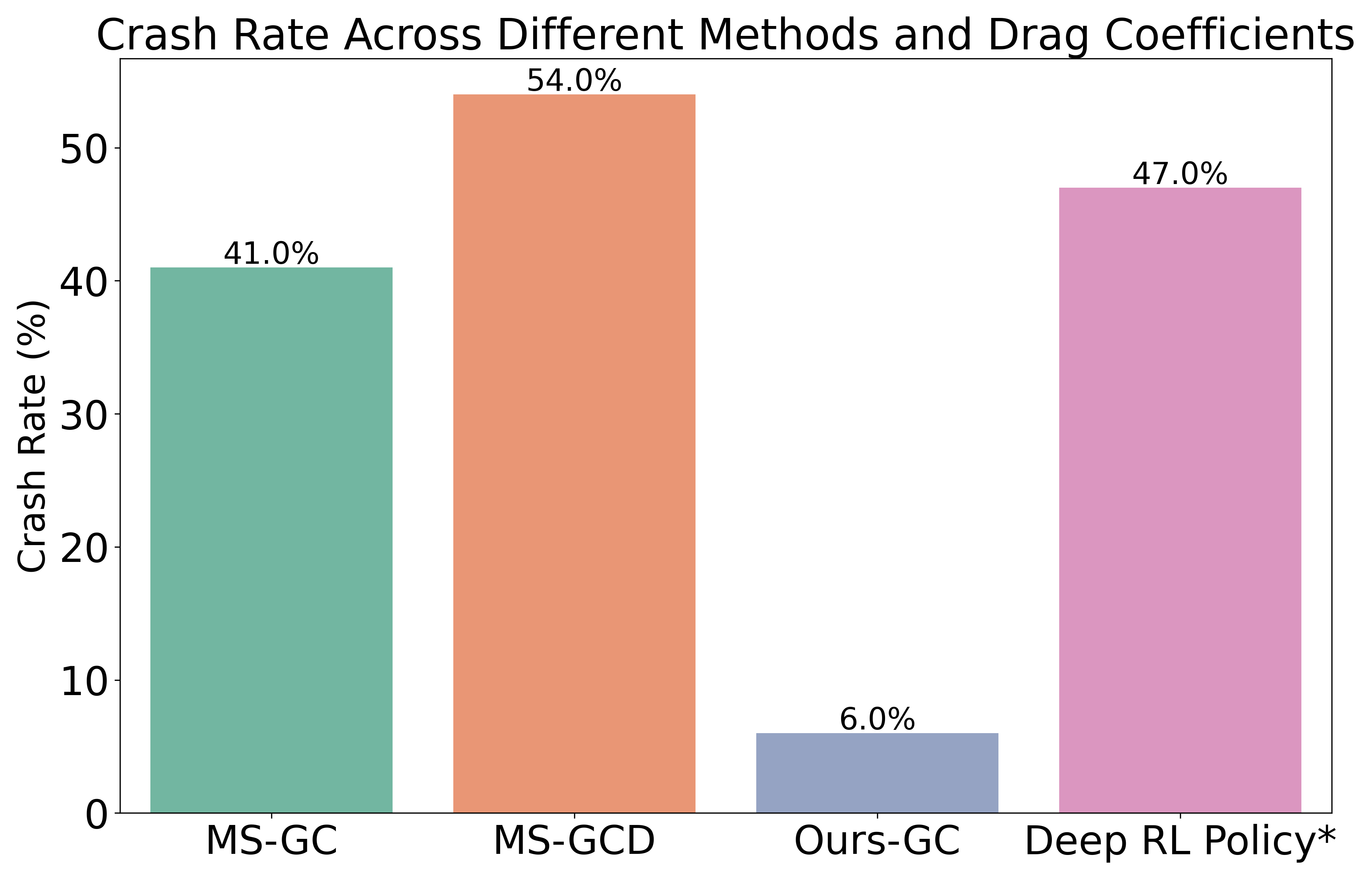}
    \caption{We report crash rates evaluated on $100$ waypoint-following tasks and average segment speed of $2 (m/s)$. The asterisk ($*$) denotes that motor noise was turned off for training and evaluation on the RL policy.}
    \vspace{-10pt}
\label{fig:hardware-sim}
\end{figure}

\textbf{Results}: After training in simulation, we evaluate each network by solving problem~\eqref{prob:drag-aware-plan} and report crash rates in Figure~\ref{fig:hardware-sim}. We define a crash for a trajectory in simulation whenever maximum position tracking over the trajectory is above $1.5 m$. Crash rate is the percentage of crashes for every $100$ evaluations. 

We note that the asterisk ($*$) on the RL policy indicates that we turned off motor noise in simulation. The RL policy has a crash rate of $47\%$ and $100\%$ with motor noise turned off and on in simulation, respectively. Further, the best trajectory in simulation had a maximum position tracking error of $38 cm$ which increases to greater than $1.5 m$ when motor delay and noise is turned on. This made any further hardware deployments for the RL policy infeasible. As seen from Figure~\ref{fig:hardware-sim}, our approach has the lowest crash rate of $6$\% compared to $41$\% for MS-GC and $54$\% for MS-GCD. 



\section{Experimental Setup for Hardware}\label{sec:exp-setup}

We evaluate zero-shot sim-to-real transfer of our proposed method from Section~\ref{sec:actuator-poly} by demonstrating experiments on a standard Crazyflie 2.0. We utilized a motion capture system that provided pose and twist measurements at 100Hz to a base station computer. The feedback controller~\eqref{eq:nonlinear-ctrl}, tuned for the Crazyflie, generated control commands in the form of collective thrust and desired attitude. The Crazyflie used onboard PID controllers and feedback from its inertial measurement unit (IMU) and the motion capture system to track these commands. Our approach as shown in Figure~\ref{fig:quad-hw} successfully planned and deployed feasible trajectories avoiding motor saturation. 



\vspace{-10pt}

\section{Conclusion}

In conclusion, we evaluate a control decomposition that mitigates motor saturation in quadrotor systems by optimizing reference trajectories with a learned cost function map. Differing from controller gain adaptation, our approach modifies trajectories and prevents uncontrolled drift, improving trajectory tracking and reducing crash rates by 49\% in aggressive maneuvers. Hardware tests on the Crazyflie validate feasibility, demonstrating real-world applicability. As future work, we aim to do a more thorough analysis of motor speeds to observe how our learned cost implicitly changes the speed profile.

\section{Acknowledgment}

We thank Fernando Cladera for their support in providing access and tools to use the computational cluster and Kashish Garg for their conversations to refactor the code~\footnote{\url{https://github.com/Nusha97/Quad-LCD}}. 

\bibliographystyle{ieeetr}
\bibliography{papers}

\end{document}